\pdfoutput=1

\documentclass[journal]{IEEEtran}
\usepackage{graphicx}
\usepackage{stackengine}
\usepackage{multirow}
\usepackage{hyperref}
\usepackage{comment}
\usepackage{amsmath,amssymb,amsthm}
\usepackage{mathtools}
\DeclarePairedDelimiter\floor{\lfloor}{\rfloor}
\usepackage{color}
\usepackage{booktabs} 
\usepackage{eso-pic}
\usepackage{pgfplots}
\usepackage{tikz}
\usepackage{standalone}
\usepackage{pgfplots}
\pgfplotsset{compat=1.14}   
    \usepgfplotslibrary{colorbrewer}
    \usepgfplotslibrary{groupplots}    
    \pgfplotsset{
        cycle list/Dark2,
        cycle multiindex* list={
            mark list*\nextlist
            Dark2\nextlist
        },
    }
\DeclareMathOperator*{\argmaxA}{arg\,max}
\DeclareMathOperator*{\argminA}{arg\,min}

\newcommand{\R}{\mathbb{R}}

\newcommand{\E}{\mathbb{E}}

\def\*#1{\mathbf{#1}}
\definecolor{bblue}{HTML}{4F81BD}
\definecolor{rred}{HTML}{C0504D}
\definecolor{ggreen}{HTML}{9BBB59}
\definecolor{ppurple}{HTML}{9F4C7C}

\newcommand{\norm}[1]{\left\lVert#1\right\rVert}

\usepackage{xcolor}
\usepackage[linesnumbered,ruled,vlined]{algorithm2e}
\ifCLASSINFOpdf
\else
\fi

\begin{document}
%
\title{Robust Character Labeling in Movie Videos: Data Resources and Self-supervised Feature Adaptation}

%

\author{Krishna~Somandepalli~\IEEEmembership{Member,~IEEE},
        Rajat~Hebbar and~Shrikanth~Narayanan~\IEEEmembership{Fellow,~IEEE}
\thanks{Department of Electrical and Computer Engineering, University of Southern California, Los Angeles,
CA, 90089; K. Somandepalli and R. Hebbar contributed equally to this work.}}

\maketitle
\begin{abstract}
Robust face clustering is a vital step in enabling computational understanding of visual character portrayal in media. Face clustering for long-form content is challenging because of variations in appearance and lack of supporting large-scale labeled data. Our work in this paper focuses on two key aspects of this problem: the lack of domain-specific training or benchmark datasets, and adapting face embeddings learned on web images to long-form content, specifically movies. First, we present a dataset of over 169,000 face tracks curated from 240 Hollywood movies with weak labels on whether a pair of face tracks belong to the same or a different character. We propose an offline algorithm based on nearest-neighbor search in the embedding space to mine hard-examples from these tracks. We then investigate triplet-loss and multiview correlation-based methods for adapting face embeddings to hard-examples. 
Our experimental results highlight the usefulness of weakly labeled data for domain-specific feature adaptation. Overall, we find that multiview correlation-based adaptation yields  more discriminative and robust face embeddings. Its performance on downstream face verification and clustering tasks is comparable to that of the state-of-the-art results in this domain. We also present the SAIL-Movie Character Benchmark corpus developed to augment existing benchmarks. It consists of racially diverse actors and provides face-quality labels for subsequent error analysis. We hope that the large-scale datasets developed in this work can further advance automatic character labeling in videos. All resources are available freely at \url{https://sail.usc.edu/~ccmi/multiface}.
\end{abstract}


%
\IEEEpeerreviewmaketitle

\section{Introduction}
%
%
%
%
\IEEEPARstart{M}{edia} is created by humans, for humans: to tell stories that educate, entertain, inform, or call us to action. When we watch a TV series or a movie, the onscreen characters shape our point of view by providing a window into the narrative. These characters advance the plot and play a vital role in effective storytelling~\cite{cohen2001defining}.
Advances in machine learning can help automatically identify the ``\textit{who, where}, and \textit{when}'' in a story and help build a computational understanding of character representations, portrayal, and behavior in media content~\cite{somandepalli2021computational}.
Such research methods and tools can also contribute to other application domains such as understanding social interactions in video \cite{vicol2018moviegraphs}, automatic video captioning \cite{rohrbach2017generating} and developing computational narratology~\cite{kim-2017}. 

Character-level analysis of media content is of interest to a variety of stakeholders -- from content creators and curators to engineers, media scholars and consumers.
Consider, for example, video streaming platforms, which are able to tailor recommendations based on the cast of characters and the settings in which they appear~\cite{el2020towards}.
Another example, particularly for content curators, is the \textit{X-ray} feature~\cite{x-ray} on Amazon's Prime Video platform.
Aimed at viewers that want to learn more about who they are watching, X-ray, among other things, identifies the cast and characters in some Prime Video content, enriching overall user experience. 
The automated tools also allow media scholars and content creators to conduct large-scale studies of media trends such as in TV shows, movies, and advertisements, using character on-screen presence to shed light on a variety of relevant topics such as diversity and inclusion in representation and portrayals~\cite{somandepalli2021computational,guha2015gender}.


A first step toward developing such technology is the ability to automatically identify the characters in the visual modality, i.e., characterize the \textit{who} in a video. 
This is a challenging task because of both the wide variability within and across individuals and their portrayals across contexts, and due to the rich variety in the design of different media forms (e.g., automatic character labeling in animated content \cite{somandepalli2017unsupervised}).
Media forms also vary with respect to duration, style, and production quality. For example, traditional feature-length movies versus micro-videos on platforms such as TikTok \cite{8845048, nie2017enhancing}. Micro-videos are typically short-form, realistic videos unlike movies which are long-form and highly produced. 
In this work, we focus on live-action movie content where a person's face is used as the primary signal of identification.
This is typically achieved through a two-step process: i) face \textit{detection} to localize the face of a person in a frame, ii) clustering the detected faces to \textit{recognize} a person irrespective of where and how they appear in the content. 
Recent advances in face detection can localize faces with near-perfect precision, even in extreme conditions of illumination and pose~\cite{deng2019retinaface}. 
Similarly, advances in learning face representations (embeddings) and rich open-source face datasets (e.g., \cite{schroff2015facenet,cao2018vggface2}) have provided powerful frameworks to identify a person by their face. 
\begin{figure*}[t!]
\centering
\includegraphics[width=0.9\textwidth]{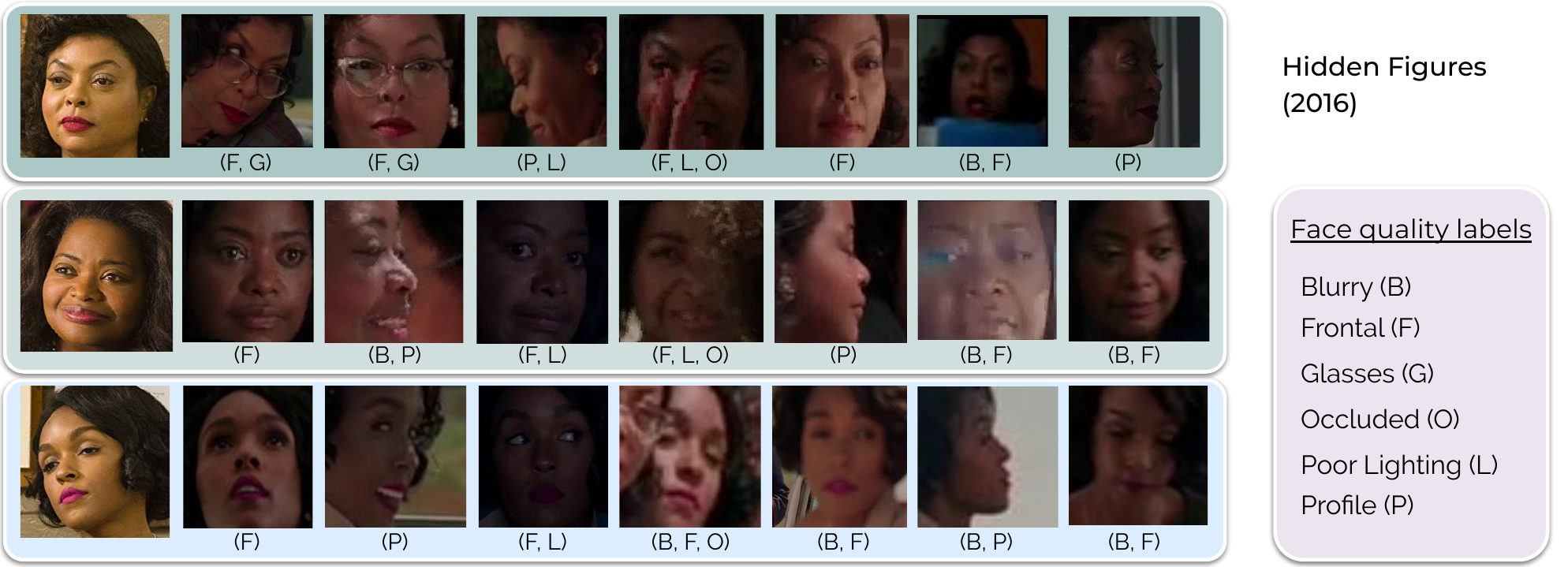}
\caption{Challenging instances of faces detected in a movie for character labeling task. The example shows the prominent characters from a 2016 Academy award winning movie \textit{Hidden Figures}. Face quality labels associated with each track are also shown to tag some of the visual distractors. The images in the first column are character label exemplars taken from the IMDb page: \url{www.imdb.com/title/tt4846340}}
\label{fig:exemplar}
\end{figure*}

However, face recognition in videos in the absence of domain-matched training data remains a challenging problem. 
We need to robustly identify the characters regardless of changes in appearance, background imagery, facial expression, size (resolution), view points (pose), illumination, partial detection (occlusion), and in some cases, age~\cite{ghaleb2015accio}. 
Figure~\ref{fig:exemplar} highlights the variability in the appearance of characters making the character labeling task challenging in the presence of visual distractors.
The task is further complicated in long-form content such as movies, where characters occur at varying frequencies and suitable exemplars of actors playing them are not always available. 
In this setup, effective face recognition not only requires face embeddings which remain robust to visual distractors, but possibly unsupervised methods such as clustering to accurately identify every character in a movie.
This is the main focus of this work with Hollywood movie videos as our exemplary application domain of interest.

Unlike photo albums or image datasets, the temporal nature of videos can be used to group time-consecutive faces detected from the same person into \textit{face tracks}, using simple heuristics based on the detection overlap~\cite{xiang2015learning} agnostic to the face identity. 
Thus, face recognition in video is generally performed at the \textit{track-level}.
Face tracking in video content such as movies---where there are typically multiple characters appearing in a scene---also offers \textit{self-supervised} means of mining pairs of faces that belong to the same person (all faces within a track) and faces that belong to different people (all faces co-occurring in a frame).
This process exploits the co-occurring nature of faces (both spatially and temporally) inherently present in videos such as movies and does not require additional supervision; hence, the term \textit{self-supervised}. 
In the context of video face clustering, several past studies have shown self-supervision frameworks to be effective in learning robust domain-specific face embeddings~\cite{zhang2016deep,sharma2019self} as well as in  improving face clustering~\cite{cinbis2011unsupervised,somandepalli2019reinforcing}.

In this paper, we use self-supervision toward two critical aspects of face clustering in long-form content: addressing the lack of domain-matched training data and adapting deep face embeddings learned from static images to movie face tracks.
First, we present a large-scale weakly labeled dataset that we curated by mining instances of spatially and temporally co-occurring face tracks in movies, called \textit{SAIL-MultiFace}.
Second, we present an offline method based on nearest-neighbor search to identify challenging cases in the weakly labeled data: that is, faces belonging to the same person which are ``far apart'' (\textit{hard-positives}) and faces of different people that are ``close together'' (\textit{hard-negatives}) in the embedding space. 
Then, to improve the overall face recognition in the movie domain using weakly labeled data, we explore triplet loss \cite{schroff2015facenet,zhang2016deep} and multiview correlation \cite{somandepalli2019multimodal} to adapt general-purpose face embeddings to the movie domain.

Finally, we present the \textit{SAIL-Movie Character Benchmark (SAIL-MCB)}, a resource developed to evaluate and compare the performance of face clustering methodologies for movies. 
For this purpose, 
we considered movie/TV videos evaluated in other studies~\cite{zhang2016deep,everingham2006hello}, and in our own recent study~\cite{somandepalli2019reinforcing}.
Finally, to ensure that this benchmark is representative and inclusive of all actors in movies, we included two other movies with a more racially diverse cast of characters.
For all six videos, we annotated nearly 10,000 face tracks with over half a million faces with character labels of both primary and secondary actors. 
We also annotated face quality labels (See an example in Figure~\ref{fig:exemplar}) to understand the performance of face recognition frameworks in the presence of visual distractors.
Our experimental results with face verification and clustering, and subsequent error analysis, demonstrate the benefit of using self-supervised adaptation techniques to improve character labeling in videos.

The rest of the paper is organized as follows: In section~\ref{sec:lit}, we discuss the relevant literature for automatic visual character labeling in movies.
We then describe the data resources we developed: SAIL-MultiFace for adaptation and the SAIL-MCB for benchmarking and evaluation in section~\ref{sec:data}; followed by a discussion of the feature adaptation methods in section~\ref{sec:methods}.
In section~\ref{sec:exp}, we consider four widely successful general-purpose face embedding frameworks to evaluate face clustering in movies with and without domain adaptation. 

\section{Related Work}\label{sec:lit}
We review here relevant existing work to contextualize the two challenges of face clustering in movies central to our work: 1. The lack of movie domain-specific data resources for training/evaluation and 2. Addressing the wide inherent variability in movie content through the ideas of self-supervision for domain adaptation.


\par{\textbf{A) Face recognition data resources.}}
One of the earliest large-scale open source datasets called Labeled Faces in the Wild (LFW), was developed in 2008~\cite{huang2008labeled}.
LFW consists of over 13,000 faces from 5,749 people detected using the Viola Jones face detector~\cite{Viola01robustreal-time}.
With the advent of deep learning methods for more robust face detection at scale, several large datasets have been created by automatically collecting face images from the Internet. 
For example, the CASIA WebFace dataset~\cite{yi2014learning} contains nearly half a million images from over 10,000 people.
Wider Face \cite{yang2016wider}, a benchmark dataset for face detection, consists of around 400,000 faces in over 32,000 search engine retrieved images.
CelebFaces Attributes (CelebA) is another large-scale resource with images from more than 20,000 celebrities.
More recently, the VGGFace2 dataset~\cite{cao2018vggface2} was released with nearly 3 million faces from over 9,000 celebrities.
Another notable effort in this space was led by Microsoft in creating MSCeleb-1M~\cite{guo2016ms} with 10 million face images from nearly 100,000 individuals. While this dataset is no longer publicly available, this effort highlights the feasibility
of curating large-scale face recognition resources. 
At the same time, such efforts also raise important ethical and privacy concerns; see~\cite{van2020ethical} for an excellent discussion.

Unlike the image domain, there have been very few fully labeled face video sources.
A notable dataset in the video domain is YouTube Faces (YTFaces,~\cite{wolf2011face}) with about 3,500 face tracks from over 1,500 different people sampled from interview recordings on YouTube.
MovieNet \cite{huang2020movienet} is another recent addition with over 1 million character labels annotated in key frames of over 1000 movies. 
The availability of such resources has led to the development of several supervised deep representations (embeddings) of faces (e.g.,~\cite{schroff2015facenet,parkhi2015deep,liu2017sphereface,deng2019arcface,shi2019probabilistic}) that have proven to be powerful in characterizing and distinguishing a person's identity.

\par{\textbf{B) Deep face embeddings.}}
Deep face embeddings are typically trained with static images mined from web search or photo albums.
Two major challenges remain in using these embeddings directly for character labeling for long-form content.
First, the unit of analysis in videos is a \textit{face track}. 
Image-level face embeddings are typically aggregated across all faces in the track using the mean operation~\cite{sharma2017simple}.
This may not be robust to dynamic changes of a person's face within a track as shown in~\cite{sharma2017simple}, resulting in unreliable track-level embeddings. 
Second, unlike web images, depictions in videos show a person's face in a wide variety of situations (backgrounds and visual distractors), particularly prevalent in long-form content such as movies~\cite{sharma2019self}, resulting in domain mismatch.
Thus image-domain embeddings do not often yield robust face representations for video content~\cite{zhang2016deep}.
One approach to address this domain mismatch is to train or adapt image-level face embeddings using labeled data from the domain-of-interest: in our case, perhaps using YTFaces dataset. 
However, for training deep learning models, such datasets are relatively small in size (3,500 face tracks) and the source videos (e.g., video interviews of celebrities with mostly frontal facing appearance) have fewer visual distractors than what is possible and expected .
Although additional domain-specific resources may be collected by manually assigning character labels to video face tracks, this process is generally time-consuming and expensive~\cite{cao2018vggface2}. 

\par{\textbf{C) Character labeling in videos.}}
Fully automatic identification of characters in video using faces has been studied for over a decade.
Some earlier works~\cite{everingham2006hello,ramanathan2014linking,sivic2009you} have explored aligning speaker names available in movie screenplays with subtitles to obtain character labels at a timestamp. 
Character identification was then framed as a matching problem of the faces detected in a movie frame with the names extracted from screenplays.
While effective in identifying at least the speaking/named characters in a video, these methods fail to scale up or generalize due to the limited access to final production screenplays as well as inaccurate time-alignment with the subtitles~\cite{cour2010talking,haurilet2016naming}.
Full-body person recognition was also studied for character identification in videos \cite{kim2018face}. 
However, movies often use close-up shots of characters and show them in different attires~\cite{rohrbach2017movie,yu2020character} throughout the movie, making the application of full-body recognition to this domain more challenging.
Audio-visual character labeling methods, particularly for movies, have been somewhat less successful (e.g.,~\cite{el2014audiovisual,kapsouras2017multimodal}); primarily because such efforts have mostly focused on identifying speaking characters (active speaker labeling~\cite{vallet2012multimodal,sharma2019toward} and fail to identify nonspeaking characters.

Using external metadata for supervised matching of characters was also explored. 
A prominent example used IMDb images for TV series as labels for character labeling~\cite{aljundi2016s}.
Such methods fail to generalize for movies because of the mismatch between the appearance of an actor's face on sources such as IMDb and their appearance in a movie (for example, the effect of makeup~\cite{kose-makeup-2015} or age~\cite{ghaleb2015accio}). 
Figure~\ref{fig:exemplar} illustrates an example of this mismatch in one of the movies in our dataset. The differences between an actor's appearance on the IMDb-curated image versus the actual appearance in the movie, are very noticeable. 
Whereas, in the present work, we focus on accurately identifying all faces belonging to a character using unsupervised methods. 
If necessary, the exemplars of the resulting clusters can be mapped to actors from casting lists with minimal manual effort.
Thus, in order to robustly represent faces in movies, we explore self-supervision for domain adaptation.
Self-supervision based research in this domain can be broadly categorized along two directions: (1) mining weakly labeled data from video content and (2) adapting face image embeddings for the domain-of-interest.
\par{\textbf{D) Self-supervision to collect weakly labeled data. }}
Local tracking of faces detected in a video acts as high precision clustering to identify faces that \textit{must} belong to the same person, and faces that \textit{cannot} belong to the same person (when multiple faces co-occur in a frame); generating \textit{must-link} and \textit{cannot-link} constraints respectively~\cite{wu2013constrained}.
Thus, without additional signals such as subtitles, metadata, or speaker labels, weakly labeled faces can be readily mined in movie content, agnostic to the character identity~\cite{somandepalli2019reinforcing}.
In the same spirit, we curate over 169,000 face tracks with weak labels from 240 movies to generate domain-matched data for feature adaptation.

While we were motivated by the success of face-tracking to mine a large number of weakly labeled tracks from movies, we used distance based methods to further identify challenging samples for downstream adaptation tasks.
Recently, two methods, track-supervised Siamese network (TSiam) and self-supervised Siamese network (SSiam) were proposed in~\cite{sharma2019self} which---besides face tracking---relied on the Euclidean distances in the embedding space to generate similar and dissimilar face tracks. 
Similarly, we propose a nearest neighbor-based approach to further segment the curated tracks into smaller ``tracklets''. Hard-positives were then identified between tracklets with maximal distance, and hard-negatives, between cannot-link tracklets with minimal distance in the embedding space.
This method is entirely offline and avoids the need for complex online hard-example mining, common in triplet-loss based systems~\cite{schroff2015facenet}.


\par{\textbf{E) Self-supervised feature adaptation. }}
The weak labels generated from must-link and cannot-link constraints have been successfully used for feature adaptation using metric learning.
Unsupervised logistic discriminative learning (ULDML~\cite{cinbis2011unsupervised}) was proposed to learn a metric such that must-link faces are closer to each other and cannot-link faces are further apart in the feature space.
A Siamese network was trained with contrastive-loss using such weakly labeled data for face verification task in~\cite{chopra2005learning} and using triplet-loss in the development of FaceNet~\cite{schroff2015facenet}.
In the context of track-level face clustering in videos, similar and dissimilar faces were used to train face embeddings using a Siamese network with contrastive loss~\cite{datta2018unsupervised}.
Triplet-loss was shown to generally perform better than contrastive loss for face recognition in videos~\cite{huo2020unique}.
An improved triplet (ImpTriplet) loss was proposed in~\cite{zhang2016deep} that performed better than the traditional triplet loss for track-level feature adaptation by additionally constraining the distance of the must-link pairs.
Thus, in our work, we evaluated ImpTriplet to adapt face embeddings using the hard-positive and hard-negative tracklets automatically gathered from movies.

Unlike contrastive/triplet loss formulation which needs a negative sample, multiview methods such as canonical correlation analysis (CCA,~\cite{hotelling1992relations}) and their deep variants~\cite{andrew2013deep} can be used to learn the shared information between a pair of positive samples; in our case, learning the shared character identity from different appearances (views) of a person. 
While deep CCA was applied effectively for face recognition~\cite{chang2018scalable} and reconstruction~\cite{zhang-2018-dcca}, multiview learning methods in general have not been explored for face clustering in videos.
Perhaps the closest in this context is using linear discriminant analysis (LDA) for face recognition in videos~\cite{pnevmatikakis2009subclass}. But, unlike CCA-based methods, LDA needs labels for all classes and just weakly-labeled data is inadequate.
In a recent work, we developed a neural-based approach called multiview correlation (MvCorr~\cite{somandepalli2019multiview}) to generalize CCA for more than two views where all we know is that a set of observations come from the same source.
In the domain of speaker recognition, we showed that MvCorr offers state-of-the-art performance for speaker clustering~\cite{somandepalli2019multiview} capturing information regarding the person's identity irrespective of the spoken utterance.
Analogous to speaker recognition, the hard-positive tracklets we extract from face tracks readily include faces of the same person in different views with respect to pose, illumination and occlusion.
Thus, for feature adaptation experiments with weakly labeled data, we explore both ImpTriplet and MvCorr frameworks.

\subsection{Benchmark datasets for movie face clustering.}
The overarching goal of face clustering in movies is to identify the characters wherever and whenever they appear throughout the video. 
In the domain of movie video analysis in particular, there are very few open-source datasets available till date to benchmark related tasks.
This is in part because feature-length movie videos are longer in duration compared to trailers and other video clips (e.g., YTFaces~\cite{wolf2011face}).
Large-scale movie datasets such as MovieNet~\cite{huang2020movienet} only provide annotations on key frames of movie videos, limiting the use of local face tracking.
Labeling characters throughout the content requires expensive manual effort and can be time intensive.

Other benchmark datasets have episodes of TV series: \textit{Buffy the vampire slayer}~\cite{everingham2006hello, zhang2016joint}, \textit{Big Bang Theory}~\cite{tapaswi2012knock} and \textit{Sherlock Holmes}~\cite{nagrani2018benedict}.
However, TV episodes are generally shorter compared to movies and do not have as much variability with respect to the appearance of characters and the backgrounds or situations in which they appear.
In the movie domain, a few widely used examples include the movies \textit{Casablanca} and \textit{American Beauty} compiled in~\cite{bojanowski2013finding}, \textit{Notting Hill}~\cite{xiao-wbslrr} and more recently ACCIO~\cite{ghaleb2015accio} which is a dataset of the \textit{Harry Potter} movies collected with a focus on learning age-invariant face representations.
In our recent work, we released labels for two other movies adding to these resources~\cite{somandepalli2019reinforcing}.
These movies and TV episodes mostly include white actors in prominent roles and are not entirely representative of the growing and desired trend of diverse casting in Hollywood (see reports~\cite{ucla-2020,seejane-2020}).
In this work, we address this limitation by developing a benchmark dataset that includes two movies with a more racially diverse casting.
Together, we hope that these resources can enable a robust evaluation of automatic character labeling methods in the movie domain.

\begin{table*}[t!]
\centering
\caption{Details of Movie Character Benchmark (SAIL-MCB) dataset. The number of characters in a given movie that were chosen to be labeled ensured a coverage of at least 99\% of the detected faces. The range of number of tracks-per-character shows that we label both prominent and minor characters based on the amount of their appearance. }\label{tab:benchmark}
\begin{tabular}{ll|rrc|ccc}
\toprule
 & Movie (year)              & \begin{tabular}[c]{@{}c@{}}{ No.}\\ { faces}\end{tabular} & \begin{tabular}[c]{@{}c@{}}{No.}\\ {tracks}\end{tabular} & \begin{tabular}[c]{@{}c@{}}{No. faces-per-track}\\ {mean $\pm$ std}\end{tabular} & \begin{tabular}[c]{@{}c@{}}{No.}\\ {characters}\end{tabular} & \begin{tabular}[c]{@{}c@{}}{No. tracks-per-character}\\ {(min, max)}\end{tabular} & \begin{tabular}[c]{@{}c@{}}{\% frontal}\\ { face tracks}\end{tabular}\\
\midrule
\textbf{ALN} & About Last Night (2014) & { 70,210}     & 1,880 & 38.2 $\pm$ 34.1       & 10             & (16, 491)     & 41.1                   \\
\textbf{BFF} & Buffy, the Vampire Slayer (2000)                 & { 47,870} & 634  & 66.3 $\pm$ 69.4      & 12             & (17, 112)     & 16.3                   \\
\textbf{DD2} & Dumb and Dumber To (2014) & { 152,908}    & 2,361  & 70.8 $\pm$ 66.3     & 10             & (12, 557)     & 24.9                   \\
\textbf{HF} & Hidden Figures (2016)   & { 101,438}    & 1,902  & 59.8 $\pm$ 61.1     & 24             & (8, 283)     & 47.3                   \\
\textbf{MT} & Maleficent (2014)               & { 53,450}     & 875   & 64.1 $\pm$ 57.9     & 10             & (14, 254)     & 28.2                   \\
\textbf{NH} & Notting Hill (1999)                & { 154,625}    & 2,121  & 79.2 $\pm$ 84.5    & 12             & (18, 585)     & 21.9                  \\
\bottomrule
\end{tabular}
\end{table*}

\section{Data Resources}\label{sec:data}
In this section we describe the two data resources we developed: (1) the SAIL Movie Character Benchmark (SAIL-MCB)
 and (2) SAIL-MultiFace: weakly labeled face tracks for training or adapting general purpose face embeddings.
All movies were purchased in house.
\subsection{SAIL-MCB: SAIL Movie Character Benchmark Dataset} 

\label{sec:benchmark_data}
We started with two widely used benchmark videos: a movie, \textit{Notting Hill}
(\textbf{NH}), and an episode from season 5 of the TV series, \textit{Buffy the Vampire Slayer}
(\textbf{BFF}. 
\textbf{NH} and \textbf{BFF} have been used to evaluate video face clustering, both online (e.g.,~\cite{kulshreshtha2020dynamic}) and offline (e.g.,~\cite{zhou2014video,bian2018video,haq2019deepstar}) methods.
Although labels are publicly available for these videos, we repeated the annotation process to improve the overall coverage of the character labels: our system detected significantly more number of faces than those in recent reports.
For example, Sharma et. al~\cite{sharma2019self} reported 39,263 faces detected for \textbf{BFF}; compare this with 8,000 more faces (See Table~\ref{tab:benchmark}, row 2) in SAIL-MCB.
We also included data from two movies which we made publicly available in our recent work~\cite{somandepalli2019reinforcing}: \textit{Dumb and Dumber To}
(\textbf{DD2}) and \textit{Maleficent}
(\textbf{MT}).
These two movies were chosen to include content produced more recently compared to \textbf{NH}.
In addition, we included \textit{About Last Night}
(\textbf{ALN}) and \textit{Hidden Figures}
(\textbf{HF}) to include a more racially diverse cast of actors.

For all six videos, we first perform face detection and local tracking using Google's API obtained with an academic license.
The face tracking used here was developed using simple heuristics based on analyzing the intersection-of-union of the successive bounding boxes of the detected faces~\cite{bugeau2008track}.
A summary of the number of faces detected and face tracks as well as the density of faces per track is shown in Table~\ref{tab:benchmark}. 
We used a face track as the unit of annotation for the movie videos. The process included two tasks: character labeling and face quality labeling.

\par{\textbf{Character labeling:}} We used the names of the actors listed on the IMDb casting page corresponding to each movie/TV series as character labels.
These labels were manually assigned to each face track by two human annotators independently and ties were resolved by a third person, thereby ensuring high annotation agreement.
The number of characters labeled was not fixed across the movies and chosen to cover at least 99\% of all faces (not face tracks) detected in each video.
This was another reason why we re-annotated \textbf{NH} and \textbf{BFF} in house.
Recent studies had only labeled for 5 characters (in \textbf{NH}~\cite{sharma2017simple}) or 6 characters (in \textbf{BFF}~\cite{sharma2017simple,sharma2019self}); compare this with 12 characters for both videos in SAIL-MCB.


As summarized in Table~\ref{tab:benchmark}, most movies only needed 10 or 12 character labels to cover 99\% of the faces detected. Hidden Figures (HF) had the largest number of characters at 24.
The number of face tracks per character varied widely within a movie. For example, in \textbf{HF}, the least frequent character has only eight face tracks while the most frequent character has 283 tracks (See Table~\ref{tab:benchmark}).
Thus, we also pick up on some characters in a minor role besides the commonly appearing characters (lead/co-leads).

\begin{figure}[t]
\centering
\scalebox{0.8}{
\begin{tikzpicture}
\begin{axis}[symbolic x coords={Profile,
Frontal,
Blurry,
Lighting,
Occluded,
Glasses},
xtick={Profile,
Frontal,
Blurry,
Lighting,
Occluded,
Glasses},  
xticklabel style={text height=0.2ex,
tick label style={rotate=45}}, 
yticklabel style={rotate=45},
ymin= 0, ymax= 60,
ytick={10, 20, 40, 60},
width=0.5\textwidth, 
bar width=15pt,
xtick style={draw=none},
ymajorgrids=true,
ylabel={Percentage of total face tracks (\%) },
] 

\addplot[ybar,fill] coordinates {
(Profile,50.22)
(Frontal,38.88)
(Blurry,11.45)
(Lighting,7.69)
(Occluded,3.59)
(Glasses,2.53)
};
\end{axis}
\end{tikzpicture}

}
\caption{Distribution of face quality labels in SAIL-MCB at the track-level. Over 50\% of the face tracks were labeled as ``Profile" which means that at least one face in the track was shown posing sideways.
}
\label{fig:face-quality}
\end{figure}
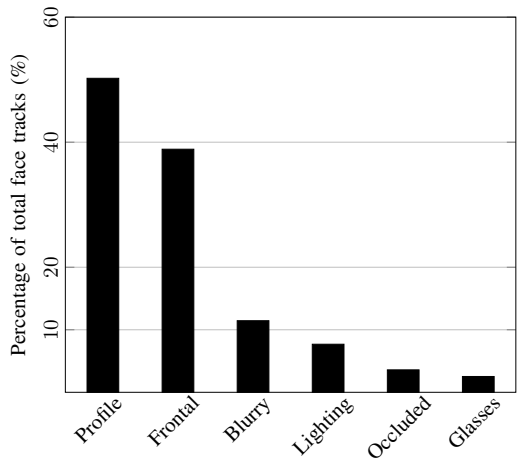

\par{\textbf{Face quality labeling:}} In order to facilitate a detailed performance and error analysis of face clustering methods on SAIL-MCB, we also obtained face quality labels for different visual distractors while labeling face tracks with their character IDs.
These qualitative labels were annotated along six dimensions at the track level: (1) whether \textit{all} faces in the track are facing the viewer (frontal:\textit{F}); (2) at least one face in the track is shown facing sideways (profile; \textit{P}); (3) at least one face in the track is blurry (\textit{B}); (4) at least one face in the track is shown wearing glasses \textit{(G)}; (5) at least one face in the track is only partially visible or occluded \textit{(O)}; and (6) at least one face in the track is poorly lit \textit{(L)}. 
These questions were presented sequentially to the annotators and they were instructed to tag with more than one label where applicable.
A few exemplars of these labels are shown in Figure~\ref{fig:exemplar}.
The distribution of the quality labels across all videos in SAIL-MCB is shown in Fig.~\ref{fig:face-quality}. In this figure, notice that the total does not sum to 100 as a single face track can have multiple distractor labels. Over 50\% of the face tracks had at least on profile face. Additionally, we found that on average, only 30\% of all face tracks were completely frontal facing in our dataset (See Table~\ref{tab:benchmark}).
These annotations along with the track-level character labels have been made publicly available.

\subsection{SAIL-MultiFace: Harvesting Weakly Labeled Face Tracks}\label{sec:harvest}
Our methodology for gathering weakly labeled data from movies includes two steps: (1) Face tracking and preprocessing to generate must-link and cannot-link pairs of face tracks and (2) hard-example mining in the embedding space to identify ``difficult'' tracklets. We call this dataset \textit{MultiFace} as it includes faces corresponding to {multiple} views of the same person in different poses or multiple settings or faces of different people in the same setting.


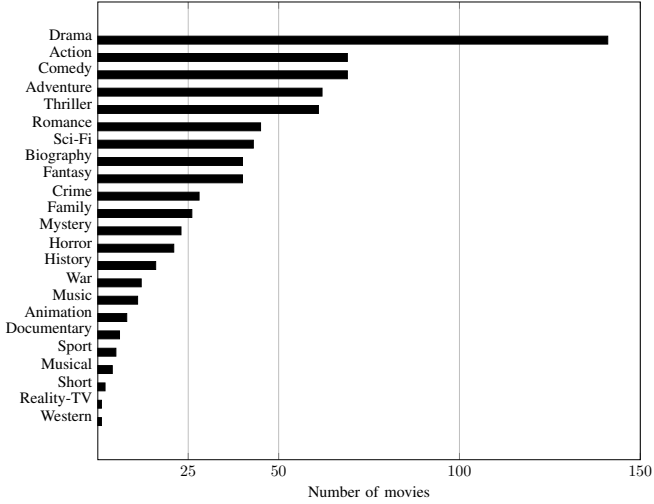
\begin{figure}[t]
\centering
\scalebox{0.6}{
\begin{tikzpicture}
\begin{axis}[symbolic y coords={ Western,
 Reality-TV,
 Short,
 Musical,
 Sport,
 Documentary,
 Animation,
 Music,
 War,
 History,
 Horror,
 Mystery,
 Family,
 Crime,
 Fantasy,
 Biography,
 Sci-Fi,
 Romance,
 Thriller,
 Adventure,
 Comedy,
 Action,
 Drama},
ytick={Western,
 Reality-TV,
 Short,
 Musical,
 Sport,
 Documentary,
 Animation,
 Music,
 War,
 History,
 Horror,
 Mystery,
 Family,
 Crime,
 Fantasy,
 Biography,
 Sci-Fi,
 Romance,
 Thriller,
 Adventure,
 Comedy,
 Action,
 Drama},  
yticklabel style={text height=0ex,
tick label style={rotate=0}}, 
xmin= 0, xmax= 150,
xtick={25, 50, 100, 150},
width=0.75\textwidth, 
bar width=5pt,
ytick style={draw=none},
xmajorgrids=true,
xlabel={Number of movies },
] 

\addplot[xbar,fill] coordinates {
(1,Western)
 (1,Reality-TV)
 (2,Short)
 (4,Musical)
 (5,Sport)
 (6,Documentary)
 (8,Animation)
 (11,Music)
 (12,War)
 (16,History)
 (21,Horror)
 (23,Mystery)
 (26,Family)
 (28,Crime)
 (40,Fantasy)
 (40,Biography)
 (43,Sci-Fi)
 (45,Romance)
 (61,Thriller)
 (62,Adventure)
 (69,Comedy)
 (69,Action)
(141,Drama)
};
\end{axis}
\end{tikzpicture}

}
\caption{Distribution of the genres for the 240 movies used for harvesting weakly labeled face tracks in SAIL-MultiFace. Movies may have multiple genres associated with them as listed on IMDb.}
\label{fig:movie-genres}
\end{figure}

The spatial and temporal co-occurrence patterns of people in a movie scene can be used to generate associations of \textit{must-link} and \textit{cannot-link} constraints i.e., faces in a track \textit{must} belong to the same person and multiple faces in a frame \textit{cannot} belong to the same person, respectively.
To gather such data in movie videos, we considered a set of 240 movies (24 fps video frame rate, 1280x720 resolution) released between 2014--2018 that were purchased in house.\newline
These movies span a wide range of genres as shown in Figure~\ref{fig:movie-genres}, providing data from different movie styles.
The movie titles and related details are provided in the supplementary methods, section I.
We performed the following preprocessing steps to mine must-link faces and cannot link tracks from each movie:
\begin{enumerate}
    \item Face detection and local tracking (as explained in the previous section).
    \item Filter out face tracks which have a duration of less than 1s, i.e., minimum of 24 faces in a track. This limits the search space of tracks for the next step as well as provides opportunities to extract instances of the same person appearing in different conditions such as pose, expression and lighting.
    \item Only retain tracks which co-occur with other tracks. We considered two face tracks to be co-occurring if they shared at least six frames (0.25s) in common. This threshold was chosen heuristically to minimize propagating errors from tracking.
\end{enumerate}
The count statistics of the total number of tracks retained at each step of the process is shown in Table~\ref{tab:harvest} for the 240 movies used for this task.
Of the tracks identified in step (2), on average, $45.1\pm21.2 \%$ of the face tracks had at least one co-occurring track.  
This statistic is consistent with the range of $35-70\%$ reported by a previous work~\cite{sharma2019self} that also mined co-occurring face tracks in movies.

It is important to note that the final number of 169,000 tracks after preprocessing only represents the {must-link} instances that provide \textit{positive samples} (faces of the same person).
In order to extract \textit{negative samples} (faces from different persons), we need to look at all cannot-link pairs associated with them. 
While every face track overlapped with at least one other track, the number of overlapping tracks varied between 1--94 with an average of $5.2\pm9.1$ overlaps per track; thus resulting in a large set of negatives to choose from. 
In such cases, past face clustering studies have emphasized the need for \textit{hard-example} mining to not only reduce the search space but also improve the robustness of face representations to visual distractors~(e.g.,~\cite{shi2019probabilistic}). 
The goal here is to identify samples belonging to the same person that appear very different (hard-positives) and samples belonging to different persons that look similar to each other (hard-negatives).
In the next section, we discuss the development of a hard-example mining approach for our use case.
\begin{table}[]
\centering
\caption{Count statistics of the tracks mined at each stage of the harvesting process. Sample size of movies = 240}\label{tab:harvest}
\begin{tabular}{l|lll}
\toprule
Statistic                                                      & \begin{tabular}[c]{@{}l@{}}Total\\ No. faces\end{tabular} & No. tracks & \begin{tabular}[c]{@{}l@{}}No. tracks/movie  \\ mean $\pm$ std.\end{tabular} \\
\midrule
Track length $\geq$ 1s & 23.2M                                                     & 335845     & 1389.9 $\pm$ 760.3                                                 \\
Co-occurring tracks                                                  & 10.2M                                                     & \textbf{169201}     & 726.2 $\pm$ 704.4    \\
\bottomrule
\end{tabular}
\end{table}

\newcommand\mycommfont[1]{\footnotesize\ttfamily\textcolor{blue}{#1}}
\SetCommentSty{mycommfont}
\SetKwInput{KwInput}{Input}                
\SetKwInput{KwOutput}{Output}              
\SetKwInput{kwInitc}{Parameters}
\SetKwInput{kwInitd}{Initialize}
\begin{algorithm*}[t]\label{algo:1}
\footnotesize
\DontPrintSemicolon
  \SetKwFunction{FKnn}{$\text{NNTracklet}_k$}
  \SetKwFunction{FHPos}{$\text{HardPositiveMining}$}
  \SetKwFunction{FHNeg}{$\text{HardNegativeMining}$}
  \SetKwFunction{Fnn}{$\text{knn}$}

  \SetKwFunction{knn}{$\text{knn}$}
  \SetKwProg{Fn}{Function}{:}{}
  
  \KwInput{The $d$-dimensional embeddings of a face-track $\*H=[\*h_1,\ldots,\*h_N]\in\mathbb{R}^{d\times N}$; Associated set of $C$ cannot-link tracks $\mathcal{C}:\{\*H^{(1)},\dotsc,\*H^{(C)}\}$}
  \KwOutput{Anchor tracklet $\*v_a$ with its hard-positive tracklets~$\mathcal{V}_p:\{\*v_p^{(1)},\dotsc,\*v_p^{(M-1)}\}$ and hard-negative tracklet~$\*v_q$}
  \kwInitc{Number of tracklets per track $M$ and nearest-neighbor parameter $k=\floor{N/M}$}
  \kwInitd{The set of must-link tracklets including hard-positives $\mathcal{V}_p=\{\}$ and cannot-link tracklets $\mathcal{V}_q=\{\}$}
  \;
  \Fn{\FHPos{$\*H$, $M$}}{
  $\*h_a = \argmaxA_{{a\in[1,N]}}\norm{\*h_a-\*h_b}^2$ \tcp*{Find anchor $\*h_a$ from pairwise distances in $\*H$}
  $\big(\*v_a\in\mathbb{R}^d,\quad{\*H'}\in\mathbb{R}^{d\times N-k}\big)\gets$\FKnn{$\*h_a$, $\*H$, $k$} \tcp*{Find anchor tracklet}
   \While{$|\mathcal{V}_p|<M$ \text{and} ${|\*H'\{:\}|}>0$}
   {
   $\*H\gets\*H'$\\
   $\*h_p=\argmaxA_{i}\norm{\*h_a - \*H\{:,i\}}^2$\tcp*{All hard-positives are mined with respect to the anchor $\*h_a$}
   $\big(\*v_p,\quad{\*H'}\big)\gets$\FKnn{$\*h_p$, ${\*H}$, $k$} \\
   $\mathcal{V}_p\gets\mathcal{V}_p\cup\{\*v_p\}$\tcp*{Find hard-positives iteratively}

   }
   \KwRet ($\*v_a, \mathcal{V}_p$)\;
  }
  \;
   \Fn{\FHNeg{$\*H$, $\mathcal{C}$}}{
    $\big(\*v_a$, $\_\big)\gets$\FHPos{$\*H$, $M$} \tcp*{Find anchor tracklet as reference}
    \While{$|\mathcal{V}_q|<MC$}
   {
   $c={|\mathcal{V}_q|}/{M}+1$\tcp*{Iterate through all the cannot-link tracks}
   	$\big(\*v'_a$, $\mathcal{V}'\big)\gets$\FHPos{$\*H^{(c)}$,$M$}\tcp*{Segment cannot-link tracks into tracklets}
        $\mathcal{V}_q\gets\mathcal{V}_q\cup\mathcal{V}'\cup\{\*v'_a\}$\tcp*{Collect tracklets from all cannot-link tracks}
   }
    $\*v_q$ = $\argminA_{\*v\in\mathcal{V}_q}\norm{\*v_a - \*v}^2$  \tcp*{Find the closest cannot-link tracklet to anchor face}
    \KwRet $\*v_q$ \;
   }
   \;
  \Fn{\FKnn{$\*q$, $\*X$, $k$}}{
    ${\*Y}\in\mathbb{R}^{d\times k}\gets$\Fnn{$\*q$, $\*X$, $k$} \tcp*{Find $k$-nearest neighbors of the query $\*q\in\mathbb{R}^d$ from the matrix~$\*X\in\mathbb{R}^{d\times N}$}
    $\*u=\frac{1}{k}\sum_{i=1}^{k}{\*Y}\{:,i\}$ \tcp*{Average nearest neigbors to form tracklet embedding}
    $\*X\gets\*X\{:,:\} \text{ -- } {\*Y}\{:,i\}$ \tcp*{Remove the nearest neighbor columns from the original matrix $\*X$}
    \KwRet ($\*u,\*X$)\;
  }
\caption{Mining Hard-positive and Hard-negative tracklets using Nearest Neighbor Search}
\end{algorithm*}
\begin{figure*}[t]
\centering
\includegraphics[height=5cm]{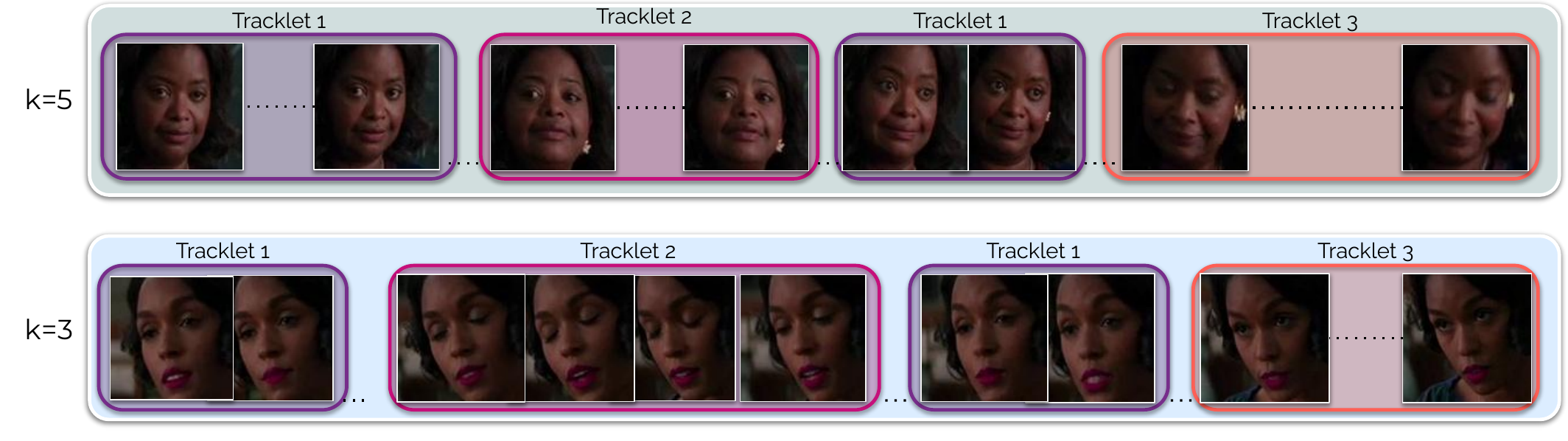}
\caption{Example of hard-positive tracklets resulting from our proposed method with the nearest neighbor parameter $k=3,5$. 
Each color indicates one tracklet.
Notice that they differ from each other with respect to face orientation (e.g., Row 1: Tracklet 1 vs. Tracklet 2) or with eyes open/closed (e.g., Row 2: Tracklet 2 vs. Tracklet 3).
A single hard-positive tracklet can be formed from faces that may not appear in a sequential order allowing us to mine \textit{harder} positives (see Tracklet 1). 
The pair of cannot-link face tracks shown here are from the movie \textit{Hidden Figures} (2016) at time $00:11:05$.}
\label{fig:tracklet}
\vspace{-0.2in}
\end{figure*}

\subsection{Mining Hard-positive and Hard-negative Tracklets}\label{sec:mining}
Hard-example mining has been studied extensively in the computer vision literature for tasks such as object recognition (e.g.,~\cite{shrivastava2016training}). 
In the context of face clustering, mining hard-positive and hard-negative \textit{faces} in an embedding space can be easily achieved by identifying the pairs of face samples that yield maximum and minimum distance respectively \cite{sharma2019self}.
However, face clustering is performed at the track-level where we typically average the embeddings of all the faces in a track to provide a robust representation.
Thus, we propose a nearest-neighbor based approach to identify the hard-positive and hard-negative \textit{tracklets}.

The pseudocode for the proposed method is detailed in Algorithm~\ref{algo:1}.
Let us assume that we have a semantic embedding space for the face tracks (e.g., VGGFace2) denoted by $\*H = [\*h_1,\dotsc,\*h_N]\in\R^{d\times N}$ with $d$-dimensional embeddings for $N$ faces in the track. 
As described in the function \FHPos{}, we segment each track into multiple tracklets to form a hard-positive set.
First, we find the pair of embeddings in $\*H$ that are maximally distant from each other and assign one of them to be the \textit{anchor} face $\*h_a$ (line 3). 
Then $k$-nearest neighbors of $\*h_a$ are accrued and averaged to form the corresponding \textit{anchor tracklet} $\*v_a$ ($\FKnn{}$, line 4)
After fixing the anchor $\*h_a$, we mine the hard-positive set $\mathcal{V}_p:\{\*h_p^{(i)}|i\in[1,M-1]\}$ iteratively (lines 5--9) as follows:\newline
(1) Find $\*h_{p}$ farthest from~$\*h_a$ from the columns remaining in $\*H$ after removing the nearest neighbors of $\*h_a$ (line 7)\newline
(2) Average the $k$-nearest neighbors of $\*h_{p}$ to get its tracklet $\*v_{p}$ (line~8)\newline
(3) Remove these nearest neighbors from the matrix $\*H$.\newline
Repeat these steps $M-1$ times to obtain the set of hard-positive tracklets $\{\*v_p^{(i)}|i\in[1,M-1]\}$ for anchor $\*v_a$.
The variable $M$ is controlled by choosing an appropriate value for the parameter $k$ in the nearest neighbor search (see \textbf{Parameters} in Algorithm~\ref{algo:1}). 
For example, if $M=5$ from a track with $N$ faces, we set $k=\floor{\frac{N}{5}}$.
This choice of $k$ ensures that all tracklets are of same length and that most faces in a track are used up.
We used the KDTree algorithm implementation in scikit-learn~\cite{scikit-learn} for nearest neighbor search.
An example of the resulting hard-positive tracklets is shown Figure~\ref{fig:tracklet}. Notice how they differ from each other with respect to face orientation and eyes closed/open.
An interesting consequence of using $k$-nearest neighbors in our approach---particularly without time-contiguous constraints---is that we can form tracklets with faces that are similar in the embedding space but may not be sequential in a face track, as highlighted in Figure~\ref{fig:tracklet}. 

Our proposed hard-negative mining is detailed in the function \FHNeg{}.
It takes two inputs: the parent track $\*H$ of anchor $\*v_a$ and the set of all $C\geq1$ cannot-link tracks for $\*H$ (see Step 3 in~\ref{sec:harvest}); denoted by $\mathcal{C}:\{\*H^{(i)}|i\in[1,C]\}$.
Each track in $\mathcal{C}$ is segmented into $M$ tracklets (line 15--16) to obtain a total of $MC$ cannot-link tracklets $\mathcal{V}_q$ (line 17).
A hard-negative is then identified as the tracklet in $\mathcal{V}_q$ that has the shortest distance to the anchor $\*v_a$ (line 18).
In Fig.~\ref{fig:tracklet}, $\mathtt{Row 1: Tracklet 3}$ was identified as the anchor and $\mathtt{Row 2: Tracklet 2}$ as its hard-negative. Notice that both tracklets show the actors with heads tilted similarly and eyes closed in a similar (dark) background.
Throughout this algorithm, we use the normalized Euclidean distance metric to estimate distances in the embedding space. Notice that both hard-positives and hard-negatives are mined with respect to an anchor which can be directly used for feature adaptation with triplet-loss based frameworks.
For all subsequent feature adaptation methods, we use the hard-positives $\{\*v_p^{(1)},\dotsc,\*v_p^{(M-1)}\}$ and the hard-negative $\*v_q$ corresponding to the anchor $\*v_a$. 

Furthermore, our proposed method is \textit{adaptive} in nature. That is, we do not need a predetermined or tuned threshold for the distance metric to identify the hard-positives and hard-negatives.
Instead, we choose the number of tracklets and let the nearest neighbor search determine the threshold.
This approach ensures that every movie in our dataset has a different \textit{minimum hard-negative distance} i.e., the closest or the most similar cannot-link pair, for which feature adaptation is necessary in the embedding space.
The distribution of the minimum hard-negative distance per movie in the SAIL-MultiFace dataset is shown supplementary methods, section II, Figure 1. 
We also discuss a few movies for which the minimum hard-negative distance is small in the embedding space. The results suggest that our proposed hard-example mining algorithm not only successfully identifies the most challenging faces to distinguish in the embedding space, but also underscores the need for feature adaptation (see supplementary methods, section II).



\section{Self-supervised Feature Adaptation}\label{sec:methods}
As discussed in Section~\ref{sec:lit}, we investigate improved triplet loss (ImpTriplet, \cite{zhang2016deep}) and multiview correlation (MvCorr,~\cite{somandepalli2019multiview}). 
Here, we review these two methods in our context of adapting general-purpose face embeddings for video face tracks using the hard-positive and hard-negative tracklets.
\subsection{ImpTriplet: Improved Triplet Loss}\label{sec:imp-triplet}
ImpTriplet~\cite{zhang2016deep} is an advanced version of the popular triplet loss formulation. For one triplet, the original triplet loss function is defined as:
\begin{align}\label{eq:orig-triplet}
    L_o = \frac{1}{2}\max{\bigg\{0,D(\*v_a, \*v_p^{(1)}) - D(\*v_a, \*v_q) + \alpha\bigg\}}
\end{align}
where $\{\*v_a, \*v_p^{(1)}\}$ and $\{\*v_a, \*v_q\}$ are the hard-positive and hard-negative tracklet pairs respectively and $\alpha$ is the distance margin (typically, $\alpha=1$).
Minimizing this loss would cause the triplet embedding to push the negative tracklet $\*v_q$ from the anchor $\*v_a$. 
However there are two key limitations in this formulation per~\cite{zhang2016deep}: (1) $\*v_q$ is pushed away only from $\*v_a$ and not both $\*v_a$ and $\*v_p^{(1)}$, and (2) the distance margin of the positive pair $\*v_a$ and $\*v_p^{(1)}$ is not specified.
ImpTriplet addresses these two issues by introducing \textit{interclass constraints} $\Phi$ and \textit{intra-class constraints} $\Psi$ respectively, as follows:
\begin{align}\label{eq:phi}
 \Phi(\*v_{a}, \*v_{p}^{(1)}, \*v_{q}) 
 & = \alpha + D(\*v_{a}, \*v_{p}^{(1)}) \\\nonumber
 & -\frac{1}{2}\big(D(\*v_{a}, \*v_{q}) + D(\*v_{p}^{(1)}, \*v_{q})\big)\\
 \Psi(\*v_{a}, \*v_{p}^{(1)}) & = D(\*v_{a},\*v_{p}^{(1)})-\hat{\alpha} 
\end{align}
Here, $\hat{\alpha}$ ensures that the anchor and positive pair lie within a margin ($\hat{\alpha}=0.1$)
Finally the ImpTriplet loss is given as:
\begin{align}\label{eq:imptriplet}
    L_s = \max{\bigg\{0, \Phi\{\*v_a, \*v_p^{(1)}, \*v_q\}\bigg\}} + \lambda\max{\bigg\{0, \Psi\{\*v_a, \*v_p^{(1)}\}\bigg\}}
\end{align}
where the parameter $\lambda$ balances the contribution of intra-class constraints in the modified triplet formulation ($\lambda=0.02$). The parameters $\hat{\alpha}$ and $\lambda$ were identified in~\cite{zhang2016deep}.

\subsection{MvCorr: Multiview Correlation}\label{sec:mvcorr}

MvCorr can successfully incorporate information from more than two views without the need for negative samples or additional labels to learn discriminative embeddings~\cite{somandepalli2019multimodal}.
For face clustering, the hard-positive tracklets containing different visual distractors can be treated as multiple views of a person's facial identity. We first discuss the loss formulation and then its application within a neural network framework.

Let $T$ be the total number of available face tracks. 
Each track is segmented into $M$ tracklets (as described in Section~\ref{sec:mining}). 
We can collect the $d$-dimensional embeddings of each tracklet as columns to form a set of $M$ hard-positive \textit{matrices} $\{\*V_a,\*V_{p}^{(1)},\dotsc,\*V_{p}^{(M-1)}\}$ with $\*V_{*}\in\mathbb{R}^{d\times T}$.
The \textit{multiview correlation} matrix $\*{\Lambda}$ is the normalized ratio of the sum of between-view covariances $\*R_b$ and sum of within-view covariances $\*R_w$ for $M$ views as follows:
\begin{eqnarray}\label{eqn:isc}
    \*{\Lambda} = \frac{1}{M-1}\frac{\*R_b}{\*R_w} = \frac{\sum_{l=1}^{M}\sum_{k=1,k\neq l}^{M}\Bar{\*V}_{l}(\Bar{\*V}_{k})^{\top}}{(M-1)\sum_{l=1}^{M}\Bar{\*V}_{l}(\Bar{\*V}_{l})^{\top}}
\end{eqnarray}
where $\Bar{\*V}_{*}=\*V_{*}-\E(\*V_{*})$ are mean-centered data matrices. The common scaling factor $(T-1)^{-1}M^{-1}$ in the ratio is omitted.
Our objective is to estimate a shared subspace, $\*W \in \mathbb{R}^{d\times d}$ such that the multiview correlation is maximized. 
Thus, the loss function can be written as:
\begin{eqnarray}\label{eqn:mcca}
    \rho^{(M)} 
    = \max_{\*W}\frac{1}{d(M-1)}\frac{\text{Tr}(\*W^{\top}\*R_b\*W)}{\text{Tr}(\*W^{\top}\*R_w\*W)}
\end{eqnarray}
where $\text{Tr}(\cdot)$ denotes the trace of a matrix. 
The subspace $\*W$ can be estimated by solving the generalized eigenvalue problem to simultaneously diagonalize $\*R_b$ and $\*R_w$. 
Hence, the MvCorr objective is the average of the ratio of eigenvalues of between-view and within-view covariances.
In other words, if $\*R_w$ is invertible, then we wish to find the a transformation matrix $\*W$ that maximizes the spectral norm of $\*R_w^{-1}\*R_b$.
Thus, maximizing the between-view variability while minimizing the within-view variability, capturing the shared information across the views.
This ratio of variances formulation is similar to that in linear discriminant analysis (LDA) but without the need for class labels.

We use neural networks to optimize MvCorr for large and complex datasets similar to our past work \cite{somandepalli2019multiview}.
Here, data from $M$ hard-positive matrices is input to $M$ corresponding `` subnetworks'' with identical architecture but no shared weights across them.
The embedding capturing the shared information across the views is the output of the last layer of the individual networks. 
The MvCorr model is trained using mini-batch SGD to minimize the loss $1-\rho^{(M)}$. 
During inference, we only need to extract embeddings from one of the  subnetworks as the last-layer activations are maximally correlated across all the  subnetworks by virtue of the loss, as demonstrated in~\cite{somandepalli2019multimodal}.

\section{Experiments and Results}\label{sec:exp}
One of the goals in this work is to empirically study whether feature adaption using weakly labeled data can lead to robust face clustering in the movie domain.
For feature adaptation, we first need to choose an embedding space.
To this end, we compare and evaluate the four of the best opensource frameworks proposed over the recent years~\cite{schroff2015facenet,cao2018vggface2,liu2017sphereface,shi2019probabilistic}.
Specifically, we setup face verification experiments to evaluate which model performs the best for the video domain, followed by feature adaptation experiments on the harvested track data SAIL-MultiFace (see Sec.~\ref{sec:harvest}).
Finally, using the SAIL-MCB dataset, we benchmark face clustering performance with and without adaptation along with a detailed error analysis using the associated face quality labels (see Fig.~\ref{fig:face-quality}).

\subsection{Face Verification for video data}\label{sec:verification}
We refer to the face embedding frameworks we tested as \textit{baseline models} since we do not additionally adapt them for the movie domain. 
We compared \textit{FaceNet} (2015,~\cite{schroff2015facenet}), \textit{SphereFace}~(2017,~\cite{liu2017sphereface}), \textit{VGGFace2} (2018,~\cite{cao2018vggface2}) and \textit{Probabilistic Face Embeddings} (\textit{PFE}, 2019,~\cite{shi2019probabilistic}).
While these frameworks have been extensively benchmarked against image datasets such as LFW~\cite{huang2008labeled}, performance comparison in the video domain is lacking.
Thus, we set up face verification experiments using two video face benchmark datasets: IJB-B~\cite{whitelam2017iarpa} 
and YTFaces~\cite{wolf2011face}.
IJB-B is commonly used for benchmarking face-verification methods in videos. 
It consists of around 77,000 faces detected from 21,000 still images and 55,000 video-frames. 
YTFaces is widely used for face recognition and verification in video, consisting of 3425 videos of approximately 1600 unique identities. While IJB-B, acquired in relatively controlled conditions helps validate if face embeddings trained on images perform well for video, YTFaces (mined from Youtube) helps evaluate their use for videos-in-the-wild.
\begin{figure*}[t!]
\centering
\scalebox{0.6}{\definecolor{Ha}{RGB}{103,163,189}
\definecolor{H1}{RGB}{92,160,56}  
\definecolor{H2}{RGB}{206,138,20}  

\begin{tikzpicture}
\begin{groupplot}[
    group style={
        group name=my plots,
        group size=3 by 1,
        ylabels at=edge left,
        xlabels at=edge bottom,
        yticklabels at=edge left,
        vertical sep=1pt
    },
    axis lines=middle,
    axis line style={->},
     x label style={at={(axis description cs:.5,-0.2)},anchor=north},
     y label style={at={(axis description cs:-0.2,.5)},rotate=90,anchor=south},
     ylabel={{\large Number of samples}},
     xmin= 0, xmax= 1.1,
     ymin= 0, ymax= 32000,
     xtick={0,0.2,0.4,0.6,0.8,1.0},
     ytick={1000, 10000, 20000, 40000},
     legend pos=north east,
    width=0.48\textwidth,
    tickpos=left,
    ytick align=inside,
    xtick align=outside
]
\nextgroupplot[title={\textbf{(a)} \stackunder{VggFace2}{(no adaptation)}}]
\addplot[ybar, area legend, bar width=10pt,fill=H1,draw opacity=0.6,opacity=0.7,rounded corners=2pt]
    table[col sep=comma, x index=0, y index=1] {must_before.dat};
\addplot[ybar, area legend, bar width=8pt,fill=H2,draw opacity=0.4,opacity=0.6,rounded corners=2pt]
    table[col sep=comma, x index=0, y index=1] {cannot_before.dat};
\nextgroupplot[xlabel={\large Normalized Euclidean distance},title={\textbf{(b)} \stackunder{Improved triplet-loss based adaptation}{(\textbf{+ImpTriplet})}}]
\addplot[ybar, area legend, bar width=10pt,fill=H1,draw opacity=0.6,opacity=0.7,rounded corners=2pt]
    table[col sep=comma, x index=0, y index=1] {must_after_triplet.dat};
\addplot[ybar, area legend, bar width=8pt,fill=H2,draw opacity=0.4,opacity=0.6,rounded corners=2pt]
    table[col sep=comma, x index=0, y index=1] {cannot_after_triplet.dat};
\legend{Hard-positives, Hard-negatives};
\nextgroupplot[title={\textbf{(c)} \stackunder{Multi-view correlation adaptation}{(\textbf{+MvCorr})}}]
\addplot[ybar, area legend, bar width=10pt,fill=H1,draw opacity=0.6,opacity=0.7,rounded corners=2pt]
    table[col sep=comma, x index=0, y index=1] {must_after_mv.dat};
\addplot[ybar, area legend, bar width=8pt,fill=H2,draw opacity=0.4,opacity=0.6,rounded corners=2pt]
    table[col sep=comma, x index=0, y index=1] {cannot_after_mv.dat};
\end{groupplot}
\end{tikzpicture}
}
\caption{Effect of feature adaption with weakly labeled data.  
Feature adaptation is expected to bring positive samples closer to each other and pull apart negative samples further in the transformed embedding space. 
Qualitative comparison of the distributions of hard-positive and hard-negative distances in SAIL-MultiFace development set shows the benefit of adaptation with (b) ImpTriplet and (c) MvCorr over the (a) original embeddings without adaptation.}
\label{fig:dist}
\end{figure*}
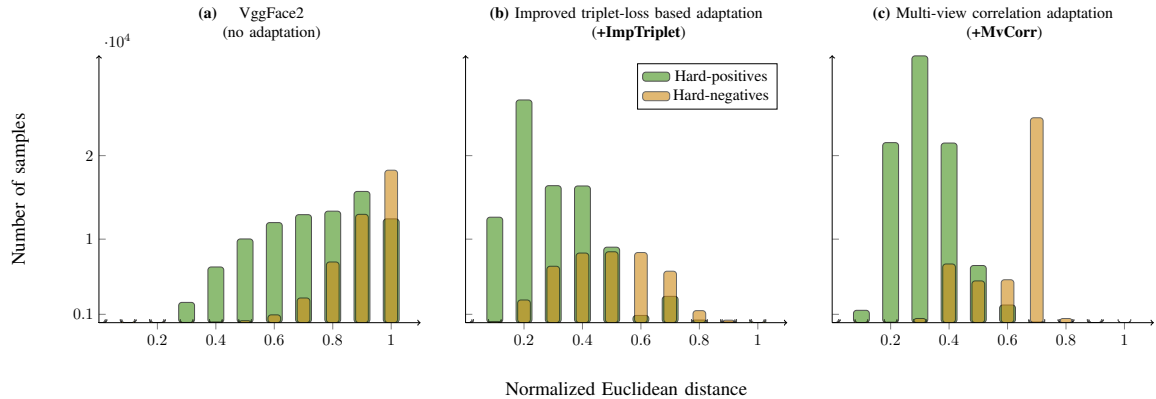
\begin{table}[t!]
\centering
\caption{Comparison of face verification performance (TPR @ 0.1 FPR $\%$) of baseline models for standard video datasets. metric}\label{tab:tpr-benchmark}
\begin{tabular}{llll}
\toprule
Model / Dataset & \begin{tabular}[c]{@{}l@{}}IJB-B\\ Aligned\end{tabular} & \begin{tabular}[c]{@{}l@{}}IJB-B\\ Unaligned\end{tabular}   & YTFaces        \\
\midrule
VGGFace2~\cite{cao2018vggface2}     & 97.0     & 97.7          & \textbf{96.6} \\
FaceNet~\cite{schroff2015facenet}   &  \textbf{98.5}        & \textbf{98.7} & 95.7          \\
SphereFace~\cite{liu2017sphereface}     & 94.4   & 52.2          & 69.2          \\
PFE~\cite{shi2019probabilistic}       & \textbf{98.3}        & 73.3          & 94.7         \\
\bottomrule
\end{tabular}
\end{table}

\textbf{Baseline models. }
\textit{FaceNet}
\cite{schroff2015facenet} 
was trained using triplet loss and Inception-v1 architecture, 
for downstream tasks such as face verification and clustering.\newline
\textit{SphereFace}~\cite{liu2017sphereface} is a metric learning method that combines the ideas of cross-entropy and angular margin loss to improve classification performance.
Notably, one of the benefits of the angular margin loss used here is its ease of training compared to triplet loss-based methods. \newline
\textit{Probabilistic face embeddings}
(PFE~\cite{shi2019probabilistic}) models different faces of the same person as multivariate Gaussian distributions where the mean captures information about the identity.
\newline
\textit{VGGFace2} is a ResNet-50 network trained on the large-scale dataset, also called as VGGFace2~\cite{cao2018vggface2} with over 9,000 face identities mined from YouTube for the task of face classification. 
In related work, \textit{VGGFace2} showed state-of-the-art face verification and clustering performance for standard datasets such as IJB-B~\cite{chang2017faceposenet}. 
For all baseline models, we used publicly available code and models pretrained on VGGFace2 dataset.
See supplementary methods, section III for preprocessing, implementation details specific to each model.

\textbf{Verification setup. }
Most face recognition methods typically perform face alignment based on facial landmarks as a pre-processing step to align faces in different poses. In movies however, these landmarks can be difficult to detect due to occlusion, pose, etc (See Fig.~\ref{fig:exemplar}). Hence, we evaluate the sensitivity of the different methods to alignment by performing two sets of verification experiments: on raw face images and \textit{aligned} face images.
For details on alignment, please see supplementary methods, section III.
Consistent with past video face experiments, we use the 1:1 verification setup described in \cite{cao2018vggface2} for IJB-B. 
For YTFaces, we create mean track-level embeddings (without alignment) and report results averaged over the 10-fold splits\footnote{YTFaces evaluation splits: \hyperlink{https://www.cs.tau.ac.il/~wolf/ytfaces/}{www.cs.tau.ac.il/wolf/ytfaces}}~\cite{wolf2011face}.
For all methods, we use the cosine distance between $l_2$-normalized embeddings as the similarity metric.

\textbf{Results. }
The ROC curves for all verification results in IJB-B with and without face alignment and YTFaces are shown in supplementary methods, section III, Figure 2. For the purposes of comparison, we choose to report TPR @ 0.1 FPR as the evaluation metric, consistent with previous reports (for example~\cite{shi2019probabilistic}). 
The results are summarized in Table~\ref{tab:tpr-benchmark}.
PFE and SphereFace are heavily reliant on face alignment and perform poorly on unaligned face images (columns 1--2, Table~\ref{tab:tpr-benchmark}). 
Without face alignment, SphereFace performed the poorest for YTFaces among the methods considered.
On the other hand, both VGGFace2 and FaceNet performed consistently well on IJB-B and YTFaces. 
While FaceNet performed slightly better than VGGFace2 in IJB-B, overall, VGGFace2 performed the best considering that face verification is more challenging for videos in-the-wild as in YTFaces.
Our findings are also comparable to the results reported elsewhere (e.g., \cite{chang2017faceposenet}).
Based on these observations, we chose VGGFace2 as the embedding space to perform movie-domain adaptation using the weakly labeled data we harvested from 240 movies (see Sec.~\ref{sec:harvest}).


\subsection{Self-supervised Feature Adaptation}
The network architecture for ImpTriplet consists of three {subnetworks}; one each for anchor $\*v_a$, hard-positive~$\*v_p^{(1)}$ and hard-negative~$\*v_q$. 
Each subnetwork is a fully-connected network (FCN) of identical architecture with shared weights across the  subnetworks.
We used publicly available code\footnote{ImpTriplet code: \hyperlink{https://github.com/manutdzou/Strong_Person_ReID_Baseline}{github.com/manutdzou/Strong\_Person\_ReID\_Baseline}} to implement the loss as described in Sec.\ref{sec:imp-triplet}.
Similarly, for MvCorr adaptation, we also use three subnetworks but without the need of hard-negatives. 
We chose $M=3$, and used the anchor $\*v_a$ and two hard-positives, $\*v_{p}^{(1)},\*v_{p}^{(2)}$ as the three multiview inputs to the network as described in~\ref{sec:mvcorr}.
We set the number of views $M$ to three in order to be comparable with the ImpTriplet adaptation which uses three tracklets per face track.
In MvCorr, the weights are not shared across the  subnetworks, unlike the ImpTriplet model.
We used our publicly released code to train MvCorr models\footnote{MvCorr code: \hyperlink{https://github.com/usc-sail/gen-dmcca}{github.com/usc-sail/gen-dmcca}}.
To choose the sub-network architecture, we explored three FCN configurations:\newline
{\footnotesize\textit{C1}: $\textbf{INP}[(512)]\rightarrow{} \textbf{FC}[512],\textbf{DO}(0.2)\rightarrow{}\textbf{FC}[256]$\\
\textit{C2}: $\textbf{INP}[(512)]\rightarrow{} \textbf{FC}[1024],\textbf{DO}(0.2)\rightarrow{}\textbf{FC}[512]$\\
\textit{C3}: $\textbf{INP}[(512)]\rightarrow{} \textbf{FC}[1024],\textbf{DO}(0.2)\rightarrow{}\textbf{FC}[512],\textbf{DO}(0.2)\rightarrow{}\textbf{FC}[256]$\\}
where \textbf{INP} = Input, \textbf{FC} = Fully connected layer with ReLu/sigmoid activation followed by batch normalization. The number of nodes in each layer is indicated inside $[\cdot]$. A dropout (\textbf{DO}) of 0.2 was added for all intermediate FC layers. Dropout was tuned over the range $\{0.1, 0.2, 0.4\}$.

\textbf{Training and model choice. }
All adaptation experiments were conducted on $169,201$ tracks in SAIL-MultiFace. 
For the training set, we used the data from 180 movies resulting in $126,435$ samples each for hard-positives and hard-negatives. 
The remaining $42,766$ samples were used as the development set.
Both adaptation networks were trained with a batch size of $1024$ using SGD (momentum=$0.9$, decay=$1e-6$) with a learning rate of $0.001$ for ImpTriplet and $0.01$ for MvCorr.
To determine model convergence, we applied early stopping criteria (stop training if the loss on the development set at the end of a training epoch did not decrease by $10^{-3}$ for $5$ consecutive epochs).
Of the \textit{C1--C3} configurations tested, we chose \textit{C2} for ImpTriplet and \textit{C1} for MvCorr as they showed the smallest loss on the development set at convergence. 
Increasing the model size beyond \textit{C3} with additional layers or changing the embedding size from $256$ to $128$ or $1024$ did not appear to further improve the loss at convergence.
All configurations showed slightly better performance with ReLu over sigmoid activation. 
All models were implemented in TensorFlow\footnote{TensorFlow 2.1: \hyperlink{https://www.tensorflow.org/api/r2.1}{tensorflow.org}} and trained on GeForce GTX 1080 Ti GPU.

\textbf{Adaptation results. }
First, we examine the distribution of the normalized Euclidean distance for all hard-positive pairs $\norm{\*v_a-\*v_p^{(1)}}^2$ and hard-negative pairs $\norm{\*v_a-\*v_q}^2$, in our development set of $60$ movies.
For hard-positives, we expect a smaller pairwise distance and the distribution to skew left, and for hard-negatives, we expect larger distances and the distribution to skew right.
The distributions for VGGFace2 embeddings without any adaptation is shown in Figure~\ref{fig:dist}a.
The distances generally skew right regardless of whether the samples were positives (similar) or negatives (dissimilar). 
The distribution of hard-positive distances is fairly uniform in the range $0.6$--$1$ showing that the face tracklets belonging to the same person can be far apart in the embedding space.
It suggests that we indeed encounter domain mismatch on direct application of VGGFace2 embeddings for face tracks in movies.
This result also highlights the effectiveness of our nearest neighbor-based hard-example mining in identifying ``difficult'' samples in the embedding space.

Next, we examine the distribution of hard-positive and hard-negative distances for ImpTriplet and MvCorr adaptation. 
As shown in Figure~\ref{fig:dist}b--c, both models skew the hard-positive distances further to the left compared to VGGFace2. 
This suggests that both methods help reduce the distance between the hard-positives as desired.
Compared to the original embeddings, ImpTriplet adaptation appears to reduce the distance between dissimilar faces (Figure~\ref{fig:dist}b) which could prove to be detrimental to downstream verification/clustering tasks.
However, MvCorr skews the distribution of hard-negative distances further to the right than ImpTriplet.
Although hard-negatives are not used in MvCorr adaptation, it appears to pull the cannot-link tracks (dissimilar faces) far apart from each other in the transformed embedding space; suggesting improved discriminability.
This is consistent with our past multiview representation learning work where multiview embeddings were able to robustly classify whether two speech segments belonged to the same person or not~\cite{somandepalli2019multimodal}. 

\begin{table}[]
\centering
\caption{Face verification performance with adaptation averaged across all videos on the SAIL-MCB benchmark dataset.}\label{tab:tpr}
\resizebox{\columnwidth}{!}{
\begin{tabular}{l|c|c|c|c}
\toprule
Metric/ Model     & {FaceNet  }                     & {VGGFace2  }                   & {+ImpTriplet  }                  & +MvCorr                       \\
\midrule
{TPR @ 0.1FPR $\%$ } & 90.2 $\pm$ 3.5 & 92.5 $\pm$ 1.7 & 91.4 $\pm$ 2.0 & \textbf{93.7 $\pm$ 1.2}\\
\bottomrule
\end{tabular}
}
\end{table}

\begin{table*}[t!]
\centering
\caption{V-measure for hierarchical agglomerative clustering (HAC) and affinity propagation (AP) with Over-clustering Index (OCI)}\label{tab:cluster}
\begin{tabular}{cl|cccccc|l}
\toprule
\multicolumn{1}{l}{} & Method      & { ALN  } & { BFF  } & { DD2  } & {  HF  } & {  MT  } & {  NH  }  & { Mean (OCI) } \\
\midrule
\multirow{3}{*}{HAC } & VGGFace2    & 83.2     & 95.3     & 82.6     & 76.7    & 88.6    & 79.5    & { 84.3 (1.0)}  \\
                     & +ImpTriplet & 81.2     & 96.6     & 83.9     & 78.0    & 89.7    & 81.1    & { 85.1 (1.0)} \\
                     & +MvCorr     & \textbf{86.8}     & \textbf{97.6}     & \textbf{85.7}     & \textbf{82.2}    & \textbf{92.1}    & \textbf{84.0}    & { \textbf{88.1} (1.0)} \\
\midrule                     
\multirow{3}{*}{AP }  & VGGFace2    & 56.3     & 76.9     & 57.1     & 65.9    & 67.5    & 59.5    & { 63.9 (5.2)}       \\
                     & +ImpTriplet & 57.4     & 77.9     & 58.8     & 67.7    & 68.9    & 60.6    & { 65.2 (6.0)}     \\
                     & +MvCorr     & \textbf{60.4}     & \textbf{84.3}     & \textbf{60.1}     & \textbf{70.1}    & \textbf{69.6}    & \textbf{62.0}    & { \textbf{67.7} (6.7)}      \\
\bottomrule                     
\end{tabular}
\end{table*}

\textbf{Face verification results. }
A possible drawback of metric-learning based adaptation is that it may transform the embedding space to optimize only for the distances between hard examples while losing the discriminability of the input embeddings.
In other words, it could overfit to the adaptation dataset.
To assess overfitting, we repeat the face verification experiments for SAIL-MCB benchmark dataset using the adapted VGGFace2 embeddings.
We generate verification pairs by exhaustively mining all combinations of face tracks using the ground-truth character labels. 
This resulted in an average of about $993,000\pm372,000$ pairs ($26 \pm 9\%$ matching pairs) across the six videos in SAIL-MCB.
We report TPR at 0.1 FPR averaged across all videos as the performance metric.
As shown in Table~\ref{tab:tpr}, the performance of ImpTriplet is comparable to that of VGGFace2. With MvCorr adaptation, we observed a small but significant ($1.2\%$) improvement in the true positive rate at 0.1 FPR. The corresponding ROC curves for face verification performance on the SAIL-MCB dataset are shown in supplementary methods, section IV, Fig. 3.



\begin{table}[t!]
\centering
\caption{Comparison of average clustering accuracy with state-of-the-art methods based on self-supervision.}\label{tab:sota}
\begin{tabular}{l|ll}
\toprule
Method/ Dataset & BFF           & NH   \\
\midrule
ULDML (2011)~\cite{cinbis2011unsupervised}           & 41.6          & 73.2 \\
HMRF (2013)~\cite{wu2013constrained}            & 50.3          & 84.4 \\
WBSLRR (2014)~\cite{xiao-wbslrr}          & 62.7          & 96.3 \\
Zhang et. al. (2016)~\cite{zhang2016joint}        & 92.1          & \textbf{99.0} \\
CP-SSC (2019)~\cite{somandepalli2019reinforcing}          & 65.2          & 54.3 \\
TSiam (2019)~\cite{sharma2019self}           & 92.5          & -    \\
SSiam (2019)~\cite{sharma2019self}           & 90.9          & -    \\
+MvCorr (ours)  & \textbf{97.7} & 96.3 \\
\bottomrule
\end{tabular}
\end{table}

\subsection{Face clustering experiments}
To test the applicability of our system for unsupervised automatic character labeling in videos, we compare the unsupervised clustering performance for VGGFace2 embeddings with and without adaptation.
For a fair comparison with past works~\cite{zhang2016deep,zhang2016joint}, we use hierarchical agglomerative clustering (HAC~\cite{scikit-learn}) and assume the number of desired clusters (unique characters) to be known.
However, in practice, the number of unique characters in a movie is often not available. 
Hence, we also experiment with affinity propagation clustering (AP~\cite{frey2007clustering}) which does not require the number of clusters before running the algorithm.
Similar to the k-medoids algorithm, AP first finds representative \textit{exemplars} to cluster all the points in the dataset. 
The exemplar count determines the number of clusters.

We evaluate the performance using several clustering metrics: homogeneity, completeness, V-measure, purity, and accuracy. 
In Table~\ref{tab:cluster}, we report the V-measure scores on the benchmark dataset. 
Performance evaluation with respect to other metrics is presented in the supplementary methods, section V.
For both HAC and AP, we achieve nearly $3\%$ improvements using ImpTriplet and $4\%$ improvement using MvCorr (See Table~\ref{tab:cluster}).
The V-measure scores for MvCorr were significantly better than VGGFace2 across all videos in our dataset\footnote{Significance testing: permutation test $n=10^5$, $p = 0.007$}. 
In contrast to HAC, the V-measure scores for AP were low, but it yields clusters which are nearly 100\% pure where multiple clusters may belong to the same character. 
We quantify this by reporting the over-clustering index~\cite{somandepalli2017unsupervised} which is the average number of clusters assigned to a character (the mean OCI across all movies is shown in Table~\ref{tab:cluster}). Mean OCI for the HAC method is $1$ because the number of clusters is provided to the clustering algorithm. 




\par{\textbf{Comparison with state-of-the-art. }}
Finally, we compare the HAC performance for clustering characters in \textbf{BFF} and \textbf{NH} to the results reported in existing works. It is important to note that although the number of characters in our dataset is greater, the comparison metric is \textit{mean} clustering accuracy, which is generally robust to these differences.
As shown in Table~\ref{tab:sota}, our proposed approach is comparable to the state-of-the-art methods for the two videos.
\begin{figure}[t]
\centering
\scalebox{0.8}{
{
\begin{tikzpicture}
    \begin{axis}[
        major x tick style = transparent,
        ybar=0.5*\pgflinewidth,
        bar width=10pt,
        ymajorgrids = true,
        ylabel = {Correctly classified face-tracks (\%)},
        symbolic x coords={Profile,Frontal,Blurry,Lighting,Occluded,Glasses},
        xtick = data,
        xticklabel style = {font=\normalsize,yshift=1.8ex},
        yticklabel style = {font=\normalsize,xshift=0.2ex},
        x tick label style={rotate=45},
        scaled y ticks = false,
        enlarge x limits=0.1,
        ymin=75,
        legend cell align=left,
        legend columns=-1,
        legend style={
                at={(1,1.0)},
                anchor=south east,
                column sep=1ex
        }
    ]
        \addplot[style={bblue,fill=bblue,mark=none}]
            coordinates {(Profile, 81.2) (Frontal, 92.2) (Blurry, 82.3) (Lighting, 91.9) (Occluded, 78.3) (Glasses, 90.2)};

        \addplot[style={ggreen,fill=ggreen,mark=none}]
            coordinates {(Profile, 91.2) (Frontal, 91.9) (Blurry, 91.1) (Lighting, 94.3) (Occluded, 88.7) (Glasses, 89.9)};

        \legend{VGGFace2 (no adaptation),+MvCorr}
    \end{axis}
\end{tikzpicture}}

}
\caption{Face quality error analysis in SAIL-MCB. 
For tracks with faces either wearing glasses (Glasses) or always facing the camera (frontal), MvCorr adaptation (+MvCorr) performed on-par with VGGFace2 pre-adaptation. 
In all other cases, +MvCorr significantly improved clustering accuracy.}
\label{fig:error-analysis}
\end{figure}
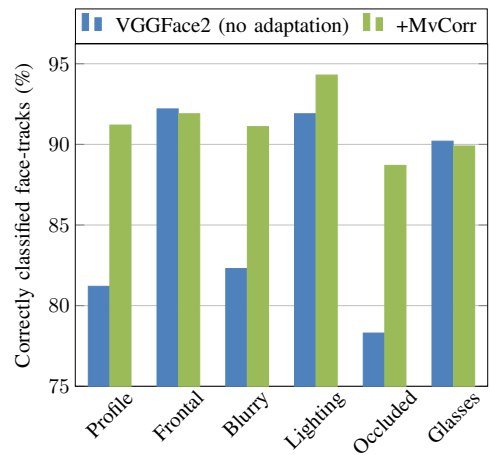

\par{\textbf{Face clustering error analysis. }}\label{sec:5}
As part of the SAIL-MCB benchmarking dataset, we also obtained face quality labels along 6 dimensions as described in Sec.~\ref{sec:benchmark_data}.
We perform error analysis of the clusters obtained using HAC along these dimensions. 
We use the percentage of total face tracks tagged with a particular quality label that were correctly classified as the metric of analysis.
To determine if a face track is correctly classified, we applied the Hungarian algorithm to the HAC output using the ground-truth character labels.
Results comparing this metric for VGGFace2 embeddings without adaptation and with MvCorr adaptation are presented in Fig.~\ref{fig:error-analysis}.
For frontal (all faces in a track facing the camera) and glasses (at least one face wearing glasses in a track), there was no significant change in performance with MvCorr adaptation. 
This is consistent with previous evaluation of VGGFace2 on datasets such as CelebA~\cite{cao2018vggface2} which showed the model to be robust to the attribute of glasses and frontal facing images.
However with all other cases (profile, blurry, poor lighting and occluded/obstructed), MvCorr adaptation significantly improved the clustering performance\footnote{Significance testing: Permutation test $n=10^4,p\leq0.01$}.
These results highlight the importance of the need for domain-matched data and subsequent adaptation to improve face clustering in movies to account for visual distractors. 

\section{Discussion}
A key component of a robust, fully unsupervised, and automatic face clustering framework for movies is the ability to discover the number of characters without using additional metadata.
In our clustering experiments, we either assumed this number to be known apriori (in HAC) or allowed for over-clustering (with AP) which requires additional heuristics or manual intervention to merge these clusters.
In this context, we want to highlight two recent and promising research directions.

An iterative merging algorithm called ball clustering~(BCL, \cite{tapaswi2019video}) was developed to jointly estimate the number of clusters as well as resolve the assignment issue of the face tracks in a video. 
However, BCL was only evaluated on TV series. While the results are encouraging with respect to clustering background/secondary characters, movies tend to have more intermittent characters than in TV series. As such, the generalizability of BCL for long-form content remains to be explored.

Recent studies of online diarization for videos in~\cite{kulshreshtha2020dynamic,Kulshreshtha2019online} proposed a shot-specific character interaction graph to incorporate constraints mined from movies. One of the benefits of online methods is that as diarization progresses new characters may be ``discovered'' as part of the process, creating new clusters.
Our future work will focus on studying these methods for face clustering in movies, particularly in a zero-shot learning framework, which has been effectively used for person re-identification in the image domain~\cite{wang2015zero}.
In the context of real-time vision applications, it is worth highlighting two recent works on video event summarization.
An alignment based method called F-DES~\cite{kumar2017f} and an AdaBoost-based approach called DELTA~\cite{kumar2018deep} were developed to identify local and global similarity patterns to identify key events from multiview videos. Our future work will consider incorporating such ideas to improve hard-example mining from weakly labeled face tracks.

Finally, although SAIL-MCB included more racially diverse movies, the face quality labels used in the error analysis only included dimensions related to visual distractors.
These labels were used to evaluate the robustness of face clustering methods.
However, these dimensions did not include demographic variables such as gender, age and race. 
We are currently working along this direction to contribute to a growing list of resources such as \textit{FairFace}~\cite{karkkainen2021fairface} which is a face image dataset with demographic variables. 
These resources can help assess the fairness of algorithms along with their robustness.


\section{Conclusion}\label{sec:conc}
In this work, we investigated robust face clustering in the movie domain using ideas of self-supervision.
First, we developed the SAIL-Movie Character Benchmark data set (SAIL-MCB) with character labels for six movie/TV videos, and SAIL-MultiFace with weakly labeled data from 240 movies, to offer domain-specific resources for feature adaptation and benchmarking.
Next, we proposed a nearest-neighbor approach to identify hard-positive and hard-negative tracklets from the must-link and cannot-link faces mined in SAIL-MultiFace. 
Finally, using these tracklets, we explored triplet-loss and multiview correlation based frameworks to adapt face embeddings learned from web images to long-form content such as movies.
Our face verification/clustering experiments and error analysis highlight the benefits of self-supervised feature adaption for robust automatic character labeling in movies.
The SAIL-MCB and SAIL-MultiFace datasets have been made publicly available.
We hope that these resources will help advance the research in understanding character portrayals in media content.


%








\bibliographystyle{IEEEtran}
\bibliography{_main}

\begin{IEEEbiography}[{\includegraphics[width=1in,height=1.25in,clip,keepaspectratio]{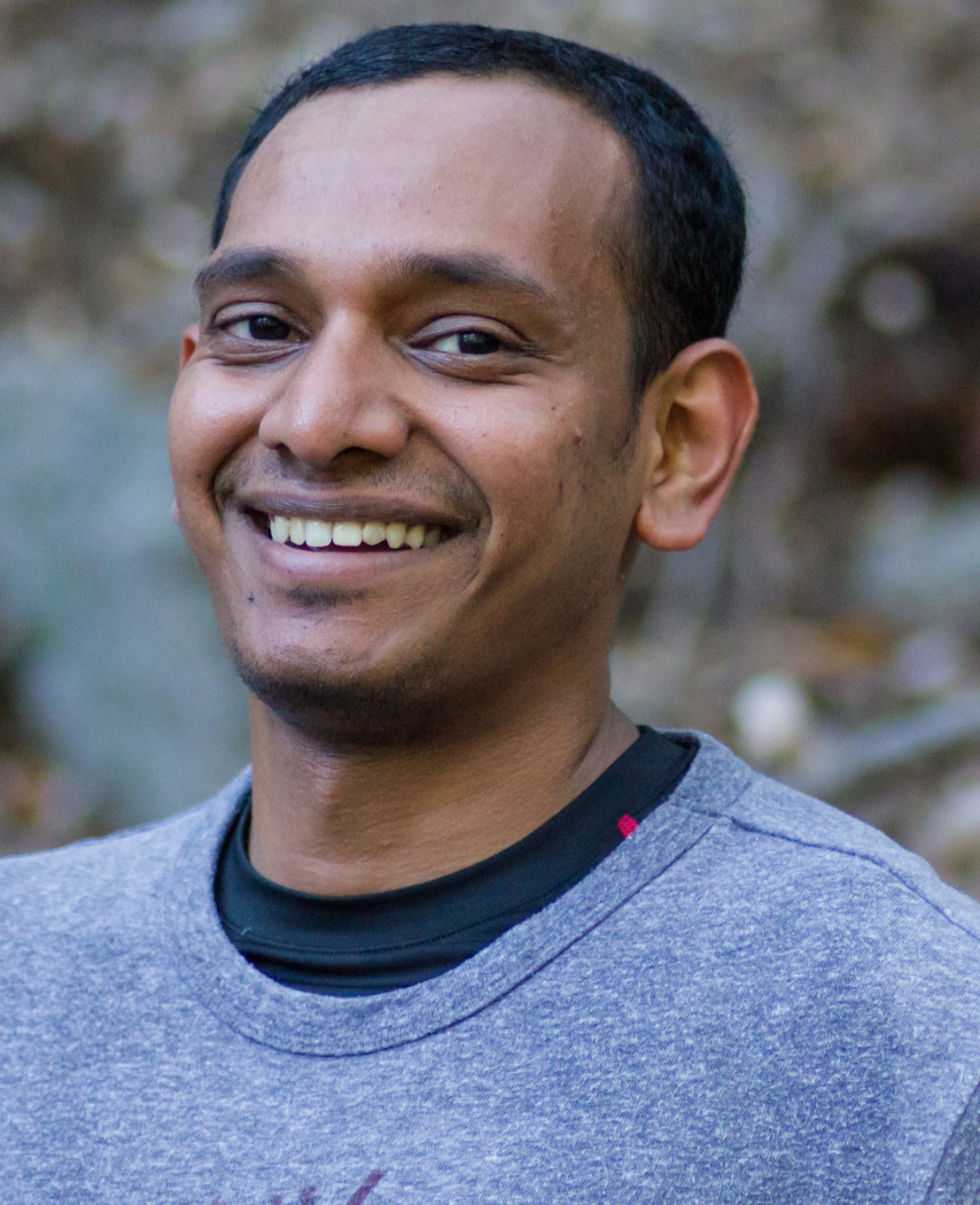}}]{Krishna Somandepalli}
Krishna Somandepalli received his PhD in Electrical and Computer Engineering from the University of Southern California, CA, USA and a Masters degree from the University of California at Santa Barbara, CA, USA in Electrical and Computer Engineering. Following his Masters degree, he worked as an assistant research scientist at NYU Langone medical Center, New York, NY, USA. He currently works at Google Research.
His research interests include multimodal signal processing, computational media understanding and developing inclusive technologies.
\end{IEEEbiography}

\begin{IEEEbiography}[{\includegraphics[width=1in,height=1.25in,clip,keepaspectratio]{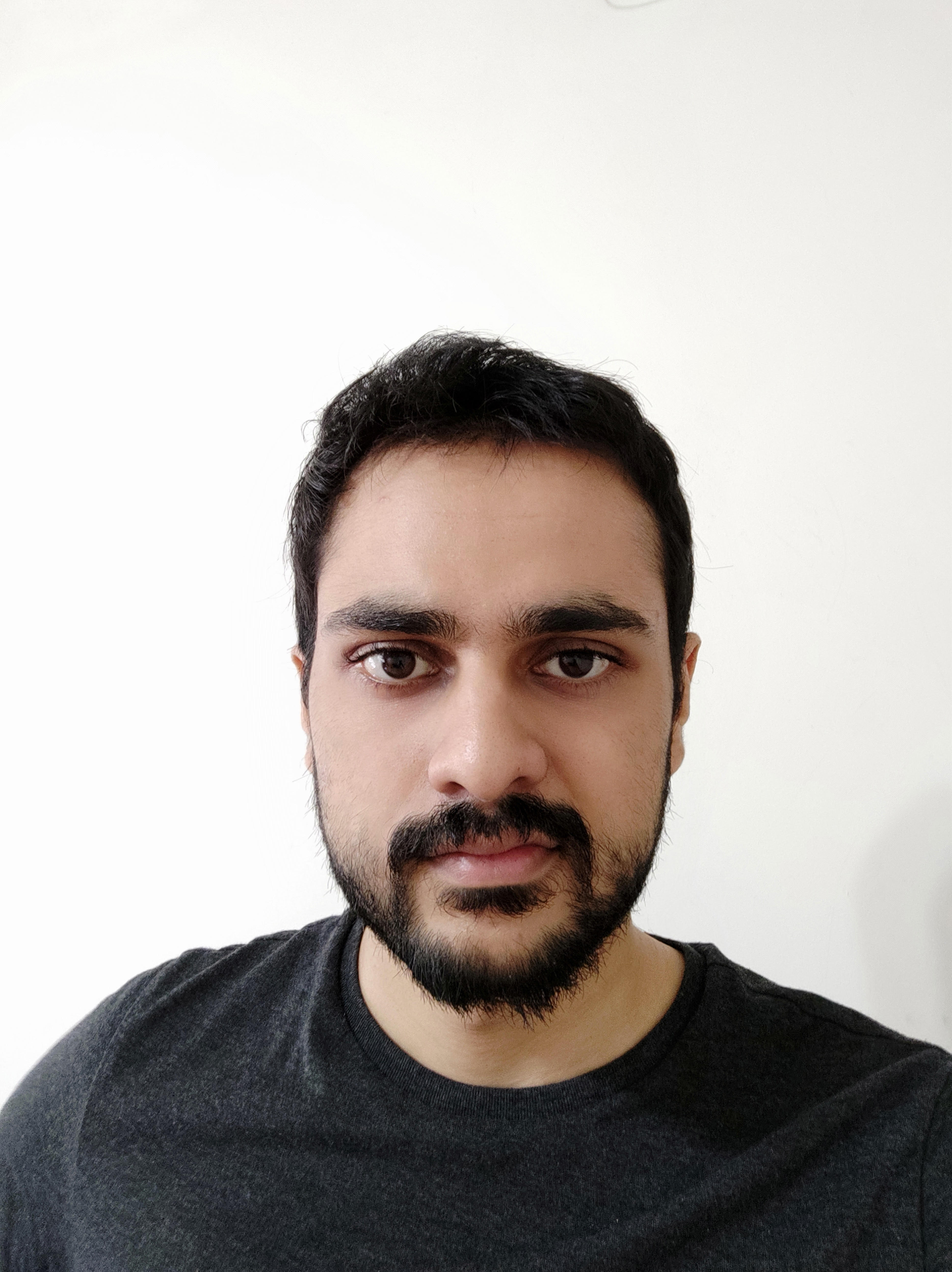}}]{Rajat Hebbar}
Rajat Hebbar is a Ph.D. student in the Electrical and Computer Engineering department at the University of Southern California, CA, USA. He obtained his Masters degree in Electrical and Computer Engineering at the University of Southern California, CA, USA, and Bachelors degree in Electronics and Communication Engineering at National Institute of Technology, Karnataka, India. His research is focused around developing machine learning-based robust speech and audio processing techniques for several challenging real-world domains, such as multimedia, wearable-device audio and multiparty meetings. 
\end{IEEEbiography}

\begin{IEEEbiography}[{\includegraphics[width=1in,height=1.25in,clip,keepaspectratio]{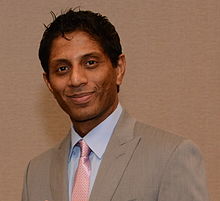}}]{Shrikanth (Shri) Narayanan}
Shri Narayanan is University Professor and Niki \& C. L. Max Nikias Chair in Engineering at the University of Southern California (USC), and holds appointments as Professor of Electrical and Computer Engineering, Computer Science, Linguistics, Psychology, Neuroscience, Otolaryngology and Pediatrics, Research Director of the Information Science Institute, and director of the Ming Hsieh Institute. Prior to USC he was with AT\&T Bell Labs and AT\&T Research from 1995-2000. At USC, he directs the Signal Analysis and Interpretation Laboratory (SAIL).   His research focuses on human-centered signal and information processing and systems modeling with an interdisciplinary emphasis on speech, audio, language, multimodal and biosignal processing and machine intelligence, and their applications with direct societal relevance. [http://sail.usc.edu]

Prof. Narayanan is a Fellow of the National Academy of Inventors, the Acoustical Society of America, IEEE, the International Speech Communication Association (ISCA), the Association for Psychological Science, the American Institute for Medical and Biological Engineering (AIMBE), and the American Association for the Advancement of Science (AAAS) and a member of Tau Beta Pi, Phi Kappa Phi, and Eta Kappa Nu. He is VP-Education, IEEE Signal Processing Society (2020-), an Editor for the Computer Speech and Language Journal and an Associate Editor for the APSIPA TRANSACTIONS ON SIGNAL AND INFORMATION PROCESSING. He was also previously Editor in Chief for IEEE JOURNAL OF SELECTED TOPICS IN SIGNAL PROCESSING and an Associate Editor of the IEEE TRANSACTIONS OF SPEECH AND AUDIO PROCESSING (2000–2004), IEEE SIGNAL PROCESSING MAGAZINE (2005–2008), IEEE TRANSACTIONS ON MULTIMEDIA (2008-2011), the IEEE TRANSACTIONS ON SIGNAL AND INFORMATION PROCESSING OVER NETWORKS (2014-2015), IEEE TRANSACTIONS ON AFFECTIVE COMPUTING (2010-2016), and the Journal of the Acoustical Society of America (2009-2017). He is a recipient of several honors including the 2015 Engineers Council’s Distinguished Educator Award, a Mellon award for mentoring excellence, the 2005 and 2009 Best Journal Paper awards from the IEEE Signal Processing Society and serving as its Distinguished Lecturer for 2010-11, a 2018 ISCA Best Journal Paper award, and serving as an ISCA Distinguished Lecturer for 2015-16 and the Willard R. Zemlin Memorial Lecturer for ASHA in 2017. Papers co-authored with his students have won awards including the 2020 Sustained Accomplishment Award and the 2014 Ten-year Technical Impact Award and 2020 Sustained Accomplishment Award from ACM ICMI and at several conferences. He has published over 900 papers and has been granted seventeen U.S. patents.
\end{IEEEbiography}
%







\end{document}